# Noisy-parallel and comparable corpora filtering methodology for the extraction of bi-lingual equivalent data at sentence level

Krzysztof Wołk

Department of Multimedia

Polish Japanese Academy of Information Technology, Koszykowa 86, 02-008 Warsaw

kwolk@pja.edu.pl

**Abstract.** Text alignment and text quality are critical to the accuracy of Machine Translation (MT) systems, some NLP tools, and any other text processing tasks requiring bilingual data. This research proposes a language independent bi-sentence filtering approach based on Polish (not a position-sensitive language) to English experiments. This cleaning approach was developed on the TED Talks corpus and also initially tested on the Wikipedia comparable corpus, but it can be used for any text domain or language pair. The proposed approach implements various heuristics for sentence comparison. Some of them leverage synonyms and semantic and structural analysis of text as additional information. Minimization of data loss was ensured. An improvement in MT system score with text processed using the tool is discussed.

## 1 Introduction

Before a parallel corpus can be used for any processing, the sentences must be aligned. Sentences in a raw corpus are sometimes misaligned, resulting in translated lines whose placement does not correspond to the text lines in the source language. Moreover, some sentences may have no corresponding translation in the corpus at all. The corpus might also contain poor or indirect translations, making the alignment difficult and quality of the data poor. Thus, it is crucial to MT system accuracy [1]. Corpora alignment must also be computationally feasible in order to be of a practical use in various applications [2].

The Polish language is a particular challenge to such tools. It is a complicated West-Slavic language with complex elements and grammatical rules. In addition, the Polish language has a large vocabulary, due to many endings and prefixes changed by word declension. These characteristics have a significant effect on the requirements for the data and the data structure itself.

In addition, English is a position-sensitive language. The syntactic order, the order of words in a sentence, plays a very significant role. The language has very limited inflection (e.g., due to the lack of declension endings). The word position in an English sentence is often the only indicator of the meaning. The sentence order follows the Subject-Verb-Object (SVO) schema, with the subject phrase preceding the predicate [9].

On the other hand, no specific word order is imposed in Polish, and the word order has little effect on the meaning of a sentence. The same thought can be expressed in several ways. For example, the sentence I just tasted a new orange juice." can be written in Polish as "Spróbowałem właśnie nowego soku pomarańczowego," or "Nowego soku pomarańczowego właśnie spróbowałem," "Właśnie spróbowałem nowego soku pomarańczowego," or "Właśnie nowego soku pomarańczowego spróbowałem." It must be noted that such differences exist in many language pairs and need to be dealt with somehow. This article proposes a language independent corpus filtering method that has been applied to Polish-English parallel corpora.

The dataset used for this research was the Translanguage English Database (TED) [3] provided by Fondazione Bruno Kessler (FBK) for IWSLT 2013[1]. The vocabulary sizes of the English and Polish texts in TED are disproportionate. There are 41,684 unique English words and 92,135 unique Polish words. This also presents a challenge for SMT systems.

Additionally, a Wikipedia comparable corpus was built and used in this research. The corpus was prepared for the needs of the IWSLT2014 Evaluation Campaign[2]. In Table 1, a fragment of the Wikipedia corpora that consists of 13 lines obtained with the Yalign tool is presented. There are many similar cases and fragments in the entire corpora that might be quite problematic during machine translation system training. First, the data is quite noisy, and the corpora contain redundant parallel lines that contain just numbers or symbols. In addition, it is easy to find improper translations, e.g., "U.S. Dept." is clearly not an accurate translation of the sentence "Na początku lat 30," which in Polish means "The beginning of the 30s". What is more, some translations are too indirect or too distinct from each other. An example of such a pair is: "In all other cases it is true." and "W przeciwnym razie alternatywa zdań jest fałszywa.," which in Polish means "Otherwise, the alternative of the sentences is false."

Despite these errors, the corpora contain good translations if the problematic data can be removed.

**Table 1.** Example of poor quality translations in the Wikipedia corpus

| Line | PL | EN |
|---|---|---|
| 1 | W przeciwnym razie alternatywa zdań jest fałszywa. | In all other cases it is true. |
| 2 | ISBN 8389795299. | ISBN 1-55164-250-6. |
| 3 | ISBN 8311068119. | ISBN 978-1-55164-311-3. |
| 4 | ISBN 8322924984. | ISBN 9780691155296. |
| 5 | ISBN 9788361182085. | ISBN 0-14-022697-4. . |
| 6 | ASN.1 (skrót od "Abstract Syntax Notation One" - abstrakcyjna notacja składniowa numer jeden) jest to standard służący do opisu struktur przeznaczonych do reprezentacji, kodowania, transmisji i dekodowania danych. | Abstract Syntax Notation One (ASN.1) is a standard and notation that describes rules and structures for representing, encoding, transmitting, and decoding data in telecommunications and computer networking. |

---

[1] http://workshop2013.iwslt.org/

[2] http://workshop2014.iwslt.org/

| | | |
|---|---|---|
| 7 | Standard ASN.1 określa jedynie składnię abstrakcyjną informacji, nie określa natomiast sposobu jej kodowania w pliku. | ASN.1 defines the abstract syntax of information but does not restrict the way the information is encoded. |
| 8 | Metody kodowania informacji podanych w składni ASN.1 zostały opisane w kolejnych standardach ITU-T/ISO. | X.683 \| ISO/IEC 8824-4 (Parameterization of ASN.1 specifications)Standards describing the ASN.1 encoding rules:* ITU-T Rec. |
| 9 | pierwszego). | One criterion . |
| 10 | problemy nierozstrzygalne. | .". |
| 11 | Jeżeli dany algorytm da się wykonać na maszynie o dostępnej mocy obliczeniowej i pamięci oraz akceptowalnym czasie, to mówi się, że jest obliczalny. | I prefer to call it merely a logical-diagram machine ... but I suppose that it could do very completely all that can be rationally expected of any logical machine". |
| 12 | temperaturę) w optymalnym zakresie. | "Algorithmic theories"... |
| 13 | Na początku lat 30. | U.S. Dept. |

## 2 Problem Analysis

Parallel sentences are a particularly valuable information source for machine translation systems, as well as for other cross-lingual information-dependent tasks. Unfortunately, such data is quite rare, especially for the Polish–English language pair. On the other hand, monolingual data for those languages is accessible in far greater quantities. We can classify the similarity of data as four main corpora types. Most rare parallel corpora can be defined as corpora that contain translation of the same document into two or more languages. Such data should be aligned at least at the sentence level. A noisy parallel corpus contains bilingual sentences that are not perfectly aligned or have poor quality translations. Nevertheless, mostly bilingual translations of a specific document should be present in it. A comparable corpus is built from non-sentence-aligned and untranslated bilingual documents, but the documents should be topic-aligned. A quasi-comparable corpus includes very heterogeneous and non-parallel bilingual documents that may or may not be topic-aligned [12].

Previous attempts to automatically compare sentences for parallel corpora were based on sentence lengths, together with vocabulary alignment [4]. Brown's method [11] was based on measuring sentence length by the number of words. Gale and Church [5] measured the number of characters in sentences. Other researchers continued exploring various methods of combining sentence length statistics with alignment of vocabularies [6, 7]. Such methods lead to the creation of noisy parallel corpora at best.

### 2.1 TED – Noisy Parallel Corpora

The Polish data in the TED talks (about 15 MB) include almost 2 million words that are not tokenized. The transcripts themselves are provided as pure text encoded i UTF-8 [8]. In addition, they are separated into sentences (one per line) and aligned in language pairs. However, some discrepancies in the text parallelism are present. These discrepancies are mainly repetitions of Polish text not included in the parallel English text.

Another problem is that the TED 2013 data is full of errors. This dataset contains spelling errors that artificially increase the dictionary size and make the statistics unreliable. A very large Polish dictionary [2] consisting of 2,532,904 different words was extracted. Then, a dictionary consisting of 92,135 unique words was created from TED 2013 data. The intersection of those two dictionaries resulted in a new dictionary containing 58,393 words. This means that 33,742 words that do not exist in Polish (e.g., due to spelling errors or named entities) were found in TED 2013. This is 36.6% of the whole TED Polish vocabulary [13].

To verify this, a manual analysis of a sample of the first 300 lines from the TED corpus was conducted. It was found that there were 4,268 words containing a total of 35 kinds of spelling errors that occurred many times. But what was found to be more problematic was that there were sentences with odd nesting, such as:

Part A, Part A, Part B, Part B., e.g.:

"Ale będę starał się udowodnić, że mimo złożoności, Ale będę starał się udowodnić, że mimo złożoności, istnieją pewne rzeczy pomagające w zrozumieniu. istnieją pewne rzeczy pomagające w zrozumieniu."

We can see that some parts (words or full phrases or even whole sentences) were duplicated. Furthermore, there are segments containing repetitions of whole sentences inside one segment. For instance:

Sentence A. Sentence A., e.g.:

"Zakumulują się u tych najbardziej pijanych i skąpych. Zakumulują się u tych najbardziej pijanych i skąpych."

or: Part A, Part B, Part B, Part C, e.g.:

"Matka może się ponownie rozmnażać, ale jak wysoką cenę płaci, przez akumulację toksyn w swoim organizmie - przez akumulację toksyn w swoim organizmie - śmierć pierwszego młodego."

The analysis identified 51 out of 300 segments that were mistaken in such a way. Overall, 10% of the sample test set contained spelling errors, and about 17% contained insertion errors. However, it must be noted that simply the first 300 lines were taken, while in the whole text there were places where more problems occurred. So, to some extent, this confirms that there were problems related to the dictionary.

In addition, there are numerous untranslated English names, words, and phrases (not translated) present in the Polish text. There are also some sentences that originate from other languages (e.g., German and French). Some translations were simply incorrect or too indirect without sufficient precision, e.g., "And I remember there sitting at my desk thinking, Well, I know this. This is a great scientific discovery." was translated into "Pamiętam, jak pomyślałem: To wyjątkowe, naukowe odkrycie." The correct translation would be "Pamiętam jak siedząc przy biurku pomyślałem, dobrze, wiem to. To jest wielkie naukowe odkrycie".

Another serious problem (especially for Statistical Machine Translation) discovered was that English sentences were translated in an improper manner.

There were four main problems:
1. Repetitions – part of the text is repeated several times after translation, i.e.:
   a. EN: Sentence A. Sentence B.
   b. PL: Translated Sentence A. Translated Sentence B. Translated Sentence B. Translated Sentence B.

2. Wrong usage of words – when one or more words used for the Polish translation slightly change the meaning of the original English sentence, i.e.:
    a. EN: We had these data a few years ago.
    b. PL (the proper meaning of the Polish sentence): We've been delivered these data a few years ago.
3. Indirect translations or usage of metaphors – when the Polish translation uses a different vocabulary in order to preserve the meaning of the original sentence, especially when the exact translation would result in a sentence that makes no sense. Many metaphors are translated this way.
4. Translations that are not precise enough – when the translated fragment does not contain all the details of the original sentence, but only its overall meaning is the same.

## 2.2 Wikipedia – Comparable Corpora

For this research, the Wikipedia corpus was extracted f from a comparable corpus generated from Wikipedia articles. It was possible to obtain about 1M topic-aligned articles from Wikipedia. The Wikipedia corpus was about 104MB in size and contained 475,470 parallel sentences. Its first version was acknowledged as permissible data for the IWSLT 2014 evaluation campaign.

The Yalign Tool [14] was used to extract parallel sentence pairs. This tool was designed to automate the parallel text mining process by finding sentences that are close translation matches to the comparable corpora. This opened up avenues for harvesting parallel corpora from sources such as translated documents and the web. What is more, Yalign is not limited to any language pair. However, creation of alignment models for the languages is necessary.

The Yalign tool was implemented using a sentence similarity metric that produces a rough estimate (a number between 0 and 1) of how likely it is for two sentences to be a translation of each other. Additionally, it uses a sequence aligner that produces an alignment that maximizes the sum of the individual (per sentence pair) similarities between two documents. Yalign's main algorithm is actually a wrapper applied prior to standard sequence alignment algorithm [14].

For the sequence alignment, Yalign uses a variation of the Needleman-Wunch algorithm [15] to find an optimal alignment between the sentences in two given documents. The algorithm has a polynomial time worst-case complexity, and it produces an optimal alignment. Unfortunately, it cannot handle alignments that cross each other or alignments from two sentences into a single one [15].

Since calculating sentence similarity is a computationally expensive operation, the implemented variation of the Needleman-Wunch algorithm uses an A* approach to explore the search space instead of using the classical dynamic programming method that requires N * M calls to the sentence similarity matrix.

After the alignment, only sentences that have a high probability of being translations are included in the final alignment. The result is filtered in order to deliver high quality alignments. To do this, a threshold value is used. If the sentence similarity metric is low enough, the pair is excluded.

For the sentence similarity metric, the algorithm uses a statistical classifier's likelihood output and adapts it into the <0,1> range.

The classifier must be trained in order to determine if a pair of sentences is an acceptable translation of each other or not. The particular classifier used in the Yalign project was a Support Vector Machine (SVM). Besides being excellent classifiers, SVMs can provide a distance to the separation hyper-plane during classification, and this distance can be easily modified using a sigmoid function to return a likelihood between 0 and 1 [16].

The use of a classifier means that the quality of the alignment depends not only on the input but also on the quality of the trained classifier.

To train the classifier, a good quality parallel dataset was necessary, as well as a dictionary with translation probability included. TED talks corpora were used for these purposes. In order to obtain a dictionary, a phrase table was trained and 1-grams were extracted from it. We used the MGIZA++ tool for word and phrase alignment. The lexical reordering was set to use the msd-bidirectional-fe method and the symmetrization method was set to grow-diag-final-and for word alignment processing [13].

Before use of a training translation model, preprocessing that included removal of long sentences (set to 80 words) had to be performed. The Moses toolkit scripts [18] were used for this purpose.

## 3 Proposed Corpora Filter

A bi-sentence filtering tool was designed to find an English translation of each Polish line in a corpus and place it in the correct place in the English file. The tool assumes that each line in a text file represents one full sentence.

Our first concept is to use the Google Translator Application Programming Interface (API) for lines for which an English translation does not exist and also for comparison between the original and translated texts. The second concept is based on web crawling, using Google Translator, Bing Translator, and Babylon translator. These can work in a parallel manner to improve performance. In addition, each translator can work in many instances. The approach can also accommodate a user-provided translation file in lieu of crowd sourcing. Any machine translation system can be used, as well, as a source of translations.

The strategy is to find a correct translation of each Polish line aided by Google Translator or another translation engine. The tool translates all lines of the Polish file (src.pl) with Google Translator and puts each line translation in an intermediate English translation file (src.trans). This intermediate translation helps in finding the correct line in the English translation file (src.en) and putting it in the correct position (Figure 1).

In reality, the actual solution is more complex. Suppose that we choose one of the English data lines as the most similar line to the specific translated line and that line has a similarity rate high enough to be accepted as the translation. This line may be more similar to the next line of src.trans, such that the similarity rate of this selected line and the next line of src.trans is higher. For example, consider the sentences and their similarity rating in Table 2.

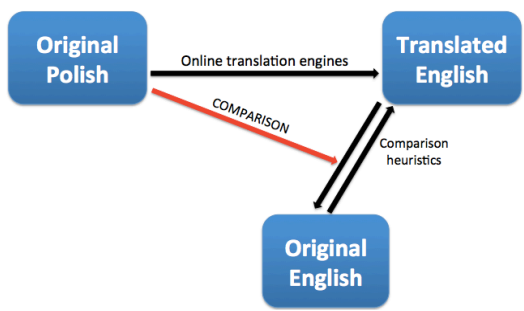

*Figure 1. Comparison Algorithm*

**Table 2.** Example Similarity Ratings

| src.trans | src.en | Sim. |
|---|---|---|
| I go to school every day. | I like going to school every day. | 0.60 |
| I go to school every day. | I do not go to school every day. | 0.70 |
| I go to school every day. | We will go tomorrow. | 0.30 |
| I don't go to school every day. | I like going to school every day. | 0.55 |
| I don't go to school every day. | I do not go to school every day. | 0.95 |
| I don't go to school every day. | We will go tomorrow. | 0.30 |

In this situation, we should select "I do not go to school every day" from src.en instead of "I don't go to school every day" from src.trans, and not "I go to school every day." So, we should consider the similarity of a selected line with the next lines of src.trans to make the best possible selection in the alignment process.

There are additional complexities that must be addressed. Comparing the src.trans lines with the src.en lines is not easy, and it becomes harder when we want to use the similarity rate to choose the correct, real-world translation. There are many strategies to compare two sentences. We can split each sentence into its words and find the number of common words in both sentences. However, this approach has some problems. For example, let us compare "It is origami." to these sentences: "The common theme what makes it origami is folding is how we create the form."; "This is origami."

With this strategy, the first sentence is more similar, because it contains all 3 words. However, it is clear that the second sentence is the correct choice. We can solve this problem by dividing the number of words in both sentences by the number of total words in the sentences. However, counting stop words in the intersection of sentences sometimes causes incorrect results. So, we remove these words before comparing two sentences.

Another problem is that sometimes we find stemmed words in sentences, for example "boy" and "boys." Despite the fact that these two words should be counted as similar in two sentences, these words are not counted with this strategy.

The next comparison problem is the word order in sentences. In Python there are other ways for comparing strings that are better than counting intersection lengths. The Python "difflib" library for string comparison contains a function that first finds matching blocks of two strings. For example, we can use difflib to find matching blocks in the strings "abxcd" and "abcd."

Difflib's "ratio" function divides the length of matching blocks by the length of two strings and returns a measure of the sequences' similarity as a float value in the range [0, 1]. This measure is 2.0*M / T, where T is the total number of elements in both sequences, and M is the number of matches. Note that this measure is 1.0 if the sequences are identical, and 0.0 if they have nothing in common. Using this function to compare strings instead of counting similar words helps to solve the problem of the similarity of "boy" and "boys." It also solves the problem of considering the position of words in sentences.

Another problem in comparing lines is synonyms. For example, consider these sentences: "I will call you tomorrow."; "I would call you tomorrow." If we want to know if these sentences are the same, we should know that "will" and "would" can be used interchangeably.

The NLTK Python module and WordNet® were used to find synonyms for each word and use these synonyms in comparing sentences. Using synonyms of each word, we created multiple sentences from each original sentence.

For example, suppose that the word "game" has the synonyms: "play", "sport", "fun", "gaming", "action", and "skittle". If we use, for example, the sentence "I do not like game.", we create the following sentences: "I do not like play."; "I do not like sport."; "I do not like fun."; "I do not like gaming."; "I do not like action."; and "I do not like skittle." We must use every word in a sentence.

Next, we should try to find the best score by comparing all these sentences instead of just comparing the main sentence. One issue is that this type of comparison takes much time, because the algorithm needs to do many comparisons for each selection.

Difflib has other functions (in SequenceMatcher and Diff class) to compare strings that are faster than the described solution, but their accuracy is worse. To overcome all these problems and obtain the best results, we should consider two criteria: the speed of the comparison function and the comparison acceptance rate.

To obtain the best results, the script provides users with the ability to specify multiple functions with multiple acceptance rates. Fast functions with lower quality results are tested first. If they can find results with a very high acceptance rate, we should accept their selection. If the acceptance rate is not sufficient, we can use slower but more accurate functions. The user can configure these rates manually and test the resulting quality to get the best results. All are well described in documentation [19].

## 4 Existing Evaluation Techniques

This section describes existing SMT evaluation techniques that highly correlate with human judgments. BLEU was developed based on a premise similar to that used for speech recognition, described in [9] as: "The closer a machine translation is to a professional human translation, the better it is." So, the BLEU metric is designed to measure how close SMT output is to that of human reference translations. It is important to note that translations, SMT or human, may differ significantly in word usage, word order, and phrase length [9].

To address these complexities, BLEU attempts to match variable length phrases between SMT output and reference translations. Weighted match averages are used to determine the translation score. [30]

A number of variations of the BLEU metric exist. However, the basic metric requires calculation of a brevity penalty $P_B$, which is calculated as follows:

$$P_B = \begin{cases} 1, & c > r \\ e^{(1-r/c)}, & c \leq r \end{cases}$$

where $r$ is the length of the reference corpus, and the candidate (reference) translation length is given by $c$. [30]

The basic BLEU metric is then determined as shown in [30]:

$$BLEU = P_B \exp\left(\sum_{n=0}^{N} w_n \log p_n\right)$$

where $w_n$ are positive weights summing to one, and the *n*-gram precision $p_n$ is calculated using *n*-grams with a maximum length of N.

There are several other important features of BLEU. First, word and phrase position within text are not evaluated by this metric. To prevent SMT systems from artificially inflating their scores by overuse of words known with high confidence, each candidate word is constrained by the word count of the corresponding reference translation. A geometric mean of individual sentence scores, with consideration of the brevity penalty, is then calculated for the entire corpus. [30]

The NIST metric was designed to improve BLEU by rewarding the translation of infrequently used words. This was intended to further prevent inflation of SMT evaluation scores by focusing on common words and high confidence translations. As a result, the NIST metric uses heavier weights for rarer words. The final NIST score is calculated using the arithmetic mean of the *n*-gram matches between SMT and reference translations. In addition, a smaller brevity penalty is used for smaller variations in phrase lengths. The reliability and quality of the NIST metric has been shown to be superior to the BLEU metric. [31]

Translation Edit Rate (TER) was designed to provide a very intuitive SMT evaluation metric, requiring less data than other techniques while avoiding the labor intensity of human evaluation. It calculates the number of edits required to make a machine translation match exactly to the closest reference translation in fluency and semantics. [18, 32]

Calculation of the TER metric is defined in [18]:

$$TER = \frac{E}{w_R}$$

where $E$ represents the minimum number of edits required for an exact match, and the average length of the reference text is given by $w_R$. Edits may include the

deletion of words, word insertion, word substitutions, as well as changes in word or phrase order. [18]

The Metric for Evaluation of Translation with Explicit Ordering (METEOR) metric is intended to take several factors that are indirect in BLEU into account more directly. Recall (the proportion of matched n-grams to total reference n-grams) is used directly in this metric. In addition, METEOR explicitly measures higher order n-grams, considers word-to-word matches, and applies arithmetic averaging for a final score. Best matches against multiple reference translations are used. [33]

The METEOR method uses a sophisticated and incremental word alignment method that starts by considering exact word-to-word matches, word stem matches, and synonym matches. Alternative word order similarities are then evaluated based on those matches. [33]

Calculation of precision is similar in the METEOR and NIST metrics. Recall is calculated at the word level. To combine the precision and recall scores, METEOR uses a harmonic mean. METEOR also rewards longer *n*-gram matches. [33]

The METEOR metric is calculated as shown in [33]:

$$METEOR = \left(\frac{10\,P\,R}{R+9\,P}\right)(1-P_M)$$

where the unigram recall and precision are given by *R* and *P*, respectively. The brevity penalty $P_M$ is determined by:

$$P_M = 0.5\left(\frac{C}{M_U}\right)$$

where $M_U$ is the number of matching unigrams, and *C* is the minimum number of phrases required to match unigrams in the SMT output with those found in the reference translations.

## 5 Comparison Experiments

Experiments were performed in order to compare the performance of the proposed method with human judgment. First, the Polish side of corpora was repaired, spell-checked, and cleaned by human translators. They were supposed to remove lines with translation into other languages or improper ones. The same was done with the usage of the filtering tool. Table 3 shows how many sentence pairs remained after the filtering, as well as the number of distinct words and their forms in the TED data.

**Table 3.** Filtering performance

|  | Sentence Pairs | PL Vocabulary | EN Vocabulary |
|---|---|---|---|
| **Baseline** | 185637 | 247409 | 118160 |
| **Human** | 181493 | 218426 | 104177 |
| **Filtering Tool** | 151288 | 215195 | 104006 |

Finally, the SMT systems were trained on the original and cleaned data to show the influence on the results. The results were compared using the BLEU, NIST, METEOR and TER metrics, which described in Section 5. The results are shown in Table 4. In these experiments, tuning was disabled because of the known MERT instability [10]. Test and development sets were taken from the IWSLT13 evaluation campaign and cleaned before usage with the help of human translators.

**Table 4.** SMT Results

|                | DATA SET | BLEU  | NIST | TER   | METEOR |
|----------------|----------|-------|------|-------|--------|
| **Baseline**   | 1        | 16.69 | 5.20 | 67.62 | 49.35  |
|                | 2        | 19.76 | 5.66 | 63.50 | 52.76  |
|                | 3        | 16.92 | 5.30 | 65.90 | 49.37  |
|                | AVG      | **17.79** | **5.38** | **65.67** | **50.49** |
| **Human**      | 1        | 17.00 | 5.33 | 66.38 | 49.90  |
|                | 2        | 20.34 | 5.78 | 62.18 | 53.55  |
|                | 3        | 17.51 | 5.39 | 64.88 | 50.14  |
|                | AVG      | **18.28** | **5.50** | **64.48** | **51.19** |
| **Filtering Tool** | 1    | 16.81 | 5.22 | 67.21 | 49.43  |
|                | 2        | 20.08 | 5.72 | 62.96 | 53.13  |
|                | 3        | 16.98 | 5.33 | 65.66 | 49.70  |
|                | AVG      | **17.94** | **5.42** | **65.27** | **50.75** |

The SMT experiments involved a number of steps. Processing of the corpora included: tokenization, cleaning, factorization, conversion to lower case, splitting, and a final cleaning after splitting. Training data was processed, and the language model was developed.

The system testing was done using the Moses open source SMT toolkit with its Experiment Management System (EMS) [23]. The SRI Language Modeling Toolkit (SRILM) [18] with an interpolated version of the Kneser-Key discounting (interpolate –unk –kndiscount) was used for 5-gram language model training. We used the MGIZA++ tool for word and phrase alignment. KenLM [24] was used to binarize the language model, with a lexical reordering set to use the msd-bidirectional-fe model.

Reordering probabilities of phrases are conditioned on lexical values of a phrase. It considers three different orientation types on source and target phrases: monotone (M), swap (S), and discontinuous (D). The bidirectional reordering model adds probabilities of possible mutual positions of source counterparts to current and subsequent phrases. Probability distribution to a foreign phrase is determined by "f" and to the English phrase by "e" [25].

MGIZA++ is a multi-threaded version of the well-known GIZA++ tool [26]. The symmetrization method was set to grow-diag-final-and for word alignment processing. First, the two-way direction alignments obtained from GIZA++ were intersected, so only the alignment points that occurred in both alignments remained. In the second phase, additional alignment points existing in their union were added. The growing step adds potential alignment points of unaligned words and neighbors. Neighborhood can be set directly to left, right, top, bottom, or diagonal (grow-diag). In the final step, alignment points between words from which at least one is unaligned are added (grow-diag-final). If the grow-diag-final-and method is used, an alignment point between two unaligned words appears. [27]

For the Wikipedia comparable corpus filtration, an initial experiment based on 1000 randomly selected bi-sentences from the corpora was conducted. The filtering tool processed the data. Most of the noisy data was removed but also some good translations were lost. Nevertheless, results are promising, and we intend to filter entire corpora in the future. It also must be noted that the filtering tool was not adjusted to this specific text domain. The results are presented in Table 5.

**Table 5.** Initial Wikipedia Filtering Results

| | |
|---|---|
| Number of sentences in base corpus | 1000 |
| Number of poor sentences in test corpus | 182 |
| Number of poor filtered sentences | 154 |
| Number of good filtered sentences | 12 |

# 6 Conclusions and Results

In general, it is a very important to create high quality parallel text corpora. Obviously, employing humans for this task is far too expensive because of the need for tremendous amounts of data to be processed. Analysis of the results of these experiments leads to the conclusion that the solution fits somewhere in between noisy parallel corpora and a corpora improved by human translators. The proposed approach is also language independent for languages with similar structure to PL or EN. The results show that the proposed method performed well in terms of the statistical machine translation evaluation. The method can also be used for improvement of noisy data, as well as comparable data. The filtering method provided a better score as compared with the baseline system, and would most likely improve the output quality other MT systems.

In summary, the bi-sentence extraction task is becoming more popular in unsupervised learning for numerous tasks. This method overcomes the disparities between English and Polish or any other West-Slavic language. It is a language independent method that can easily be adjusted to a new environment, and it only requires the adjustment of initial parameters. The experiments show that the method performs well. The corpora used here increased the MT quality in a wide text domain, the TED Talks. We can assume that even small differences can make a positive influence on real life translation scenarios. From a practical point of view, the method requires neither expensive training nor language-specific grammatical resources, while producing satisfactory results.

## 7 Acknowledgements

This work is supported by the European Community from the European Social Fund within the Interkadra project UDA-POKL-04.01.01-00-014/10-00 and Eu-Bridge 7th FR EU project (grant agreement n°287658).